\documentclass[twoside,10pt,journal,cspaper,compsoc]{IEEEtran}
\usepackage{amssymb}
\usepackage{amsmath}
\usepackage{bbm}
\usepackage{dsfont}

\usepackage{graphicx}

\usepackage{url}

\usepackage{color}

\usepackage{mdwtab}
\usepackage{caption}

\ifCLASSOPTIONcompsoc
\else
\fi
\ifCLASSINFOpdf
\else
\fi
\begin{document}
%
\title{Learning to Diversify via Weighted Kernels for Classifier Ensemble}
%
%
%
%

\author{Xu-Cheng Yin,~\IEEEmembership{Member,~IEEE,}
        Chun Yang,
        and Hong-Wei Hao~
\IEEEcompsocitemizethanks{\IEEEcompsocthanksitem X.-C. Yin is with the Department
of Computer Science and Technology, School of Computer and Communication Engineering, University of Science and Technology Beijing, Beijing 100083, China.\protect\\
E-mail: xuchengyin@ustb.edu.cn
\IEEEcompsocthanksitem C. Yang is with the Department
of Computer Science and Technology, School of Computer and Communication Engineering, University of Science and Technology Beijing, Beijing 100083, China.
\IEEEcompsocthanksitem  H.-W. Hao is with the Institute of Automation, Chinese
Academy of Sciences, Beijing 100190, China.}
\thanks{}}

%
%

\markboth{}%
{X.-C. Yin \MakeLowercase{\textit{et al.}}: Learning to Diversify via Weighted Kernels for Classifier Ensemble}
%


\IEEEcompsoctitleabstractindextext{%
\begin{abstract}
Classifier ensemble generally should combine diverse component classifiers.
However, it is difficult to give a definitive connection between diversity measure and ensemble accuracy.
Given a list of available component classifiers, how to adaptively and diversely ensemble classifiers becomes a big challenge in the literature.
In this paper, we argue that diversity, not direct diversity on samples but adaptive diversity with data,
is highly correlated to ensemble accuracy,
and we propose a novel technology for classifier ensemble, \textbf{learning to diversify},
which learns to adaptively combine classifiers by considering both accuracy and diversity.
Specifically, our approach, \textbf{L}earning \textbf{TO} \textbf{D}iversify via \textbf{W}eighted \textbf{K}ernels (\textbf{L2DWK}),
performs classifier combination by optimizing a direct but simple criterion:
maximizing ensemble accuracy and adaptive diversity simultaneously by minimizing a convex loss function.
Given a measure formulation, the diversity is calculated with weighted kernels (i.e., the diversity is measured on
the component classifiers' outputs which are kernelled and weighted), and the kernel weights are automatically learned.
We minimize this loss function by estimating the kernel weights in conjunction with the classifier weights, and propose a self-training algorithm for conducting this convex optimization procedure iteratively.
Extensive experiments on a variety of $32$ UCI classification benchmark datasets show that the proposed approach consistently
outperforms state-of-the-art ensembles such as Bagging, AdaBoost, Random Forests, Gasen, Regularized Selective Ensemble, and Ensemble Pruning via Semi-Definite Programming.

\end{abstract}

}

\maketitle

\IEEEdisplaynotcompsoctitleabstractindextext

%
\IEEEpeerreviewmaketitle

\section{Introduction}
%
%

%
%
%
%
\IEEEPARstart{T}{here} are many famous ensembles (e.g., Bagging~\cite{Breiman96}, Boosting~\cite{Freund96, Freund97},
Stacking~\cite{Wolpert92}, Random Forests~\cite{Breiman01} and neural network ensembles~\cite{Hansen90, Zhou02})
and also many recent ensemble methods~\cite{Li2013, Ma2013},
many of which have been widely applied in numerous real-world applications.
Some previous researches show that the performance of a classifier ensemble relies on not only the accuracy
but also the diversity of base classifiers~\cite{Hansen90,Opitz96,Krogh97,Zhou12}.
Generally speaking, ensemble of diverse classifiers can allow us to get higher accuracy, which is often not achievable by a single model.
Consequently, how to diversely combine classifiers plays an important role and becomes a main topic in ensemble classifier.

Given a diversity measure formulation, most conventional diversity-based ensemble methods combine classifiers by calculating and evaluating this fixed measure directly,
e.g. \cite{Li09, Martinez06, Martinez09}.
Several diversity measures are extensively acknowledged in classifier ensemble, e.g., \emph{Disagreement}, \emph{Q-Statistics}, \emph{Double Fault}, and \emph{Kappa}~\cite{Skalak96, Shipp02, Kuncheva03}.
 Obviously, the diversity is useful, and an ensemble should combine diverse component classifiers for improving accuracy; in an extreme case, ensembling all same classifiers has no any improvement. However, how to adaptively use diversity and suitably optimize combination is still unclear and challenging in the literature.
Up to now, many research evidences showed that accuracy did not monotonically increase with diversity using various diversity measures, and seemingly verified that there was no clear correlation between diversity and accuracy in ensembles~\cite{Ruta01,Kuncheva03,Saitta06}.
Some researchers even argued that in practice it was not possible to define and use a diversity measure that is clearly linked to the ensemble accuracy~\cite{Saitta06}.
\textbf{The point is ``how to use the useful diversity"}.
Currently, this seems like a ``fast knot" in ensemble learning.

From a different view to conventional diversity-based ensembles,
firstly, we argue that the diversity (called as \textbf{adaptive diversity with data}) should be strongly correlated to the accuracy, not directly measured on the training data but adaptively on the latent data distribution. Currently, given data samples, all existing diversity measures are calculated on the validation set without considering data noise and measure space, although there are a few related descriptions.
For example, Ruta and Gabrys observed that certain diversity measures (e.g., Q statistics) performed well on artificial data, but was inadequate for realistic datasets~\cite{Ruta01}.
Secondly, 
we also believe that diversity-based ensembles, i.e., ensembles with both accuracy and (adaptive) diversity, can improve performance of the final classification system.

 Consequently, we introduce a new term in classifier ensemble, \textbf{learning to diversify}~\footnote{In information retrieval, given a list of documents or web pages, learning to diversify is to learn to rank and present a diverse set of results by considering both relevance and diversity~\cite{Radlinski08,Radlinski09}. Similarly, here given a list of available component classifiers, learning to diversify in classifier ensemble is to learn to select and combine a diverse set of classifiers by considering both accuracy and diversity.}, which should learn to adaptively select and combine a subset of classifiers by considering both accuracy and diversity, given a list of available component classifiers.
There are two main essential tasks of this \textbf{learning to diversify} strategy:
(1) First and obviously, same as diversity-based ensembles, the finally selected classifiers in an ensemble should be more diverse than the general combination strategy, e.g., averaging;
(2) second but more importantly, according to different data under various situations, \textbf{learning to diversify} should adaptively evaluate the diversity with the given formulation, i.e., adaptive diversity with data.

Specifically, within a linear classifier ensemble framework, we propose a unified learning-to-diversity approach, Learning TO Diversify via Weighted Kernels (\textbf{L2DWK}).
For the first task, we propose to learn \textbf{classifier weights} by optimizing a direct but simple criterion:
maximizing the accuracy and the diversity simultaneously. Moreover, we formulate this procedure in a convex minimization problem.
For the second task, different from the conventional methods where the diversity is directly calculated on the component classifiers' outputs from the validation samples,
we seek adaptive diversity with data in the view of measure space (with kerneling) and data noise (with weighting).
The diversity in our approach is adaptively evaluated with a set of weights (called as \textbf{kernel weights}) on the samples' outputs within a kernel, i.e., the classifiers' outputs are converted into another space with weighted kernels, and the \textbf{kernel weights} are learned automatically according to the contribution of validation samples.
Furthermore, we combine these two tasks into a unified framework with a self-training algorithm. In this algorithm, first, given the kernel and weights (\textbf{kernel weights}), the \textbf{classifier weights} for combination are learned by minimizing a convex loss function, which maximizes accuracy and diversity of the ensemble simultaneously. Then, the \textbf{kernel weights} are automatically updated with a dynamically damped learning trick.
This two-step optimization is then repeated until convergence or
the maximum number of iterations is exceeded.
We argue that this diversity measure via weighted kernels is adaptive and robust for noise data in different situations, and thereby classifier ensemble based on \textbf{learning to diversify} can produce a subset of important and diverse classifiers.
To some extent, we hopefully arrive the above point: ``\textbf{to use the useful diversify}" with an adaptively learning strategy.

%
Our approach is extensively evaluated and verified on a large set of $32$ typical UCI classification datasets, and consistently outperforms state-of-the-art ensembles such as Bagging, AdaBoost, Random Forests, Gasen~\cite{Zhou02}, Regularized Selective Ensemble~\cite{Li09}, and Ensemble Pruning via Semi-Definite Programming~\cite{Zhang06}.

The rest of the paper is organized as follows.
Related work is presented in Section~\ref{sec:relatedwork}. Section~\ref{sec:notation} describes some notations used in this paper.
Our proposed method, Learning TO Diversify via Weighted Kernels (L2DWK), is presented in Section~\ref{sec:l2d} in detail.
Section~\ref{sec:exp} shows extensive experimental results on $32$ UCI benchmark datasets.
Finally, some conclusions are drawn in Section~\ref{sec:conclusion}.

\section{Related Work}
\label{sec:relatedwork}

Classifier ensemble can be mainly divided into two categories. The first one aims at learning multiple classifiers at the feature level, where multiple classifiers are trained and combined in the learning process, e.g., Boosting~\cite{Freund96} and Bagging~\cite{Breiman96}. The second tries to combine classifiers at the output level, where the results of multiple available classifiers are combined to solve the targeted problem, e.g., multiple classifier systems (classifier combination).
For example, in most learning application systems, there are a lot of decision experts derived from various homogeneous or heterogeneous resources. Given a list of these learners, one applicable solution is to merge all these decisions, e.g., the merging framework in the Waston DeepQA~\cite{Gondek2012}.
In this paper, we focus on the second one. Namely, given multiple classifiers (available or sequently learned, homogeneous or heterogeneous), classifier ensemble is learned by combining intelligently these component classifiers.
Our Learning TO Diversify via Weighted Kernels is a proper technique to combine various available classifiers in such general learning systems.

From the view of ensemble pruning, Zhou divided related methods into three categories: ordering-based pruning, clustering-based pruning, and optimization-based pruning approaches~\cite{Zhou12}. More specifically, optimization-based pruning methods formulate the ensemble pruning problem as an optimization problem that aims to find the subset of available component classifiers which maximizes or minimizes an objective related to the generalization ability of the final ensemble.
Our L2DWK ensemble falls into the optimization-based pruning ensemble methods.
Main techniques of optimization-based pruning include heuristic optimization~\cite{Zhou02}, mathematical programming (e.g., ensemble pruning via semi-definite programming~\cite{Zhang06}, and selective ensemble under regularization framework with quadratic programming~\cite{Li09}), and probabilistic pruning methods~\cite{Chen09}.

\subsection{Diversity-Based Classifier Ensembles}

As described above, diversity should be a necessary condition for high generalization ability of classifier ensemble. However, there exists many challenges in diversity-based classifier ensemble, e.g., diversity measurement, effectiveness analysis, and ensemble optimization. In the literature, most researchers try to solve these issues in two ways.

First, researchers want to give proper formulations for measuring ensemble diversity. Actually, several diversity measures have been proposed, where \emph{Disagreement}~\cite{Skalak96}, \emph{Double Fault}~\cite{Giacinto2000}, \emph{Kappa}~\cite{sim2005}, and \emph{Difficult}~\cite{Hansen90} are common ones. Specifically, Kuncheva and Whitaker compared 10 popular diversity measures (4 pair-wise and 6 non-pair-wise ones) on a variety of benchmark datasets~\cite{Kuncheva03}. They found that in most cases, non-pair-wise methods gained a slightly higher accuracy than pair-wise ones. However, non-pair-wise measures have a much higher complexity. In most optimization-based ensemble methods, pair-wise diversity measures are chosen. 

Second, based on diversity measures, many researchers want to find proper strategies for combining classifiers. Numerous diversity-based ensembles have also been proposed, most of which are related to ensemble pruning.
Margineantu et al. proposed a Kappa pruning approach, which aimed at maximizing the pair-wise diversity among the selected component classifiers~\cite{Margineantu97}.
Martinez et al. sorted component classifiers by diversity and selected the top $20\%-40\%$ classifiers for combining~\cite{Martinez09}.
Li et al. tried to present a theoretical study on the effect of diversity on the generalization performance of voting in the PAC-learning framework for classifier ensemble. Following this analysis, they also proposed the diversity regularized ensemble pruning method~\cite{Li12}.
Trawinski et al. used Genetic Algorithm (GA) to search the best classifier subset by a linear combination of accuracy and diversity~\cite{Trawinski09}.
Yin et al. proposed a heuristic learning approach with diversity and sparsity to learn classifiers' weights and combine multiple classifiers~\cite{Yin2012iconip,Yin2013neucom}.
The major problem with the above methods is that when it comes to optimizing some criteria of the selected subset, they all resort to greedy search and may get stuck in a local optima.

To get the global optima, several ensemble pruning methods with mathematical programming optimization have been proposed.
Zhang et al.~\cite{Zhang06} proposed a Semi-Definite Programming (SDP) approach, which formulates the ensemble pruning problem as a quadratic integer programming problem and solve it by a semi-definite programming solution technique. In their method, the size of the selected subset after pruning should be known in advance.
Li et al.~\cite{Li09} proposed Regularized Selective Ensemble (RSE) approach, which formulates the ensemble pruning problem as a quadratic programming problem, and learns classifier weights that have the optimal accuracy-diversity trade-off.
By introducing slack variables, the solution of RSE involves computing the weights of validation samples, and it turns out to be effective. 

However, some researchers argue that it is difficult to give a definitive connection between diversity measure and ensemble accuracy.
For example,
Ruta and Gabrys~\cite{Ruta01} studied the relationship between majority voting errors and diversity measures operating on binary classifier outputs, and found that some measures were consistently correlated with majority vote error, but there was no clear correlation between diversity and accuracy.
Kuncheva and Whitaker pointed out that it was difficult to give a definitive connection between the diversity measures and the improvement of the accuracy, and the use of diversity measures for enhancing the design of classifier ensembles was still an open question~\cite{Kuncheva03}.
 Saitta analyzed and experimented a numerous diversity measures, and concluded that all these measures have only a very loose relation with the expected accuracy of the classifier~\cite{Saitta06}.
 Tang et al. also showed from experiments that the relationship between diversity and margins of an ensemble was not so obvious, e.g., the minimum margin of an ensemble was not monotonically increasing with respect to diversity~\cite{Tang06}.

Currently, most existing diversity measures are evaluated on the validation set without considering data noise and measure space with given samples.
In this paper, by introducing kernel methods, we expand existing diversity measures with weighted kernels, and propose a new diversity-based ensemble method, Learning TO Diversify via Weighted Kernels, which learns to adaptively combine classifiers by considering both accuracy and diversity.

\subsection{Kernel Methods}
Kernel methods~\cite{Muller2001, Campbell2002, Hofmann2008} approach the problem by mapping the data into a high dimensional feature space, where each coordinate corresponds to one feature of the data items. 
In that space, a variety of methods can be used to find relations in the data. Since the mapping can be quite general (e.g., not necessarily linear), the relations found in this way are accordingly very general.

According to different situations and applications, many kernel functions are proposed, e.g., Linear, Gaussian, Polynomial,  Bayesian~\cite{ZhangDJ11}, Wave~\cite{AubrySC11} and Wavelet kernels~\cite{Wei12}, where Linear, Gaussian and Polynomial kernels are three common ones. Their definitions are as follows.

The \textbf{Linear Kernel} is the simplest kernel function. It is given by the inner product $\langle x,y\rangle$ of variable pairs plus an optional constant $c \in \mathds{R}$,
\begin{equation}
\label{eq:linearkernal}
\Bbbk_l(x,y) = x^Ty+c
\end{equation}
The \textbf{Gaussian Kernel} is an example of the radial basis function kernel ($\sigma \in \mathds{R}$),
\begin{equation}
\label{eq:gausskernal}
\Bbbk_g(x,y) = exp({-\frac{(x-y)^2}{2\sigma^2}})
\end{equation}
The \textbf{Polynomial Kernel} is a non-stationary kernel,
\begin{equation}
\label{eq:polykernal}
\Bbbk_p(x,y) = (x^Ty+c)^d
\end{equation}
where $c \in \mathds{R}$ and $d \in \mathds{N}$. 
In this paper, we use these three common kernel functions to expand the diversity measures.

In recent years, kernel methods are widely used in feature selection.
For example,
Guyon et al. proposed SVM-RFE (Recursive Feature Elimination) to eliminate features from the full set sequentially, and used it to select genes for cancer classification~\cite{Guyon02}.
Lin and Zhang proposed COSSO (COmponent Selection and Smoothing Operator) to select features in smoothing spline regression by breaking up the regularization term into components on individual dimensions~ \cite{Lin06}.
For the above kernel-based feature selection methods, all possible feature combinations must be explored in order to find the optimal solution. This is usually intractable. Hence, most feature selection methods use heuristics which make it unclear how good the solution is.

One alternative solution for this problem is the weighted kernel method, which parameterizes the kernel with a weight on each feature~\cite{Rasmussen06}.
An example of
the weighted kernel with two $d$-dimension variables $x$ and $y$ is
$$
\Bbbk(x,y)=\exp(-\sum_{i=1}^d \beta_i \|x_i^T-y_i\|^2)
$$
where $\beta$ are the weights and the Gaussian kernel is performed in Gaussian processes~\cite{Rasmussen06}. Similarly, some researchers also used SVMs for feature selection~\cite{GrandvaletC02, MangasarianK07}.
More specifically, Keerthi et al. proposed an efficient algorithm that alternates between learning an SVM and optimizing the feature weights~\cite{KeerthiSC06}.
Varma and Babu proposed a projected gradient method with $l_1$ regularization for multiple kernel learning (with weighted kernels) and reported encouraged results in a variety of benchmark vision and UCI databases~\cite{Varma09}.

Inspired by feature selection with weighted kernels,
considering data noise and measure space for different data
under various situations, we introduce learning to diversify via weighted kernels to
evaluate the diversity measure, and then select and ensemble available classifiers with both accuracy and diversity.

\section{Notations}
\label{sec:notation}
Let the training dataset be $Tr = \{x_1,x_2,...,x_N\}$, where $y_i$ is the output of sample $x_i$, and all the outputs are in $C$ classes $\{\omega_1,\omega_2,...,\omega_C\}$.
The base classifiers $H =\{h_1,h_2,...,h_L\}$ of ensemble are trained on the training set, and an output of a base classifier $h_j$ on sample $x_i$ is $h_j(x_i)$.
Given each base classifier $h_j$, together with its weight $\textrm{w}_j$, we define the vector of \textbf{classifier weights} as ${\bf w}=[\textrm{w}_1,\textrm{w}_2,...,\textrm{w}_L]$,
where $\sum_{j=1}^L{\textrm{w}_j}=1,~\textrm{w}_j \ge 0$.
In this paper, we focus on the linear combination of classifiers.
By taking a weighted vote among the base classifiers and choosing the class label receives the largest weighted vote, the ensemble $H$ classifies sample $x_i$ as $H(x_i)$.
Instead of the original output, the oracle output ${\bf O}$ of the ensemble is often used for ensemble optimization, which is a $N \times L$ matrix, and its element is
\begin{equation}\label{eq:Oracle-Output}
  O_{ij}=\left\{\begin{array}{rl}
  1& h_j(x_i)=y_i\\
  -1& h_j(x_i)\neq y_i\\
  \end{array}
  \right.
\end{equation}

The main notations used in this paper are summarized as follows:

1) $N$: number of total samples.

2) $L$: number of total base classifiers.

3) $O_{ij}$: the oracle output of the base classifier $h_j$ on sample $x_i$.

4) ${\bf O}_j$: the oracle output of the base classifier $h_j$ on the whole sample set, i.e., ${\bf O}_j=[O_{1j},O_{2j},..,O_{Nj}]^T$.

5) $m_i$: the margin of sample $x_i$, and
\begin{equation}\label{eq:Margin}
\footnotesize
\begin{array}{rl}
  m_{i} = \sum_{h_j(x_i)=y_i}\textrm{w}_j-\sum_{h_j(x_i)\neq y_i}\textrm{w}_j
        = \sum_{j=1}^L{\textrm{w}_jO_{ij}}.\\
\end{array}
\end{equation}

6) ${\bf P}$: the average ``accuracy" of classifiers on the training set. ${\bf P}=[P_1,P_2,...,P_L]^T$, where $P_j$ is the accuracy of the $j^{th}$ classifier.
Here, in order to simplify the formula derivation, the accuracy is defined as
\begin{equation}\label{eq:Accuracy}
{\bf P} = \frac{1}{N}{\bf 1}_{N\times 1}^T{\bf O},
\end{equation}
instead of the traditional one, i.e., $${\bf P}=\frac{1}{2N}{\bf 1}_{N\times 1}^T({\bf O}+{\bf 1}_{N\times L}).$$

7) $div({\bf w})$:  the diversity of ensemble with classifier weights ${\bf w}$.
If we use the pairwise diversity, $div()$ can be calculated by averaging with
\begin{equation}\label{eq:diversity-1}
\begin{array}{rl}
    div({\bf w}) =& {\bf w}^T{\bf D}{\bf w}  \\
    {\bf D} = & f_D({\bf O}^T{\bf O},{\bf 1}_{N\times 1}^T{\bf O}) \\
\end{array}
\end{equation}
where ${\bf D}$ is the diversity matrix of base classifiers, and ${\bf D}$ has a functional relationship $f_D()$ with ${\bf O}^T{\bf O}$ and ${\bf 1}^T{\bf O}$.
For example, the diversity matrixes with two common pairwise diversity measures (\emph{Disagreement}~\cite{Skalak96} and \emph{Double Fault}~\cite{Giacinto2000}) are shown below.
The diversity matrix with \emph{Disagreement} (\textbf{dis}) is
\begin{equation}\label{eq:Dis}
    {\bf D_{dis}} = \frac{1}{2N}(N{\bf 1}_{L \times L}- {\bf O}^T{\bf O}).
\end{equation}
The diversity matrix with \emph{Double Fault} (\textbf{df}) is calculated with
\begin{equation}\label{eq:Df}
\footnotesize
\begin{array}{rl}
   {\bf D_{df}} = &\frac{1}{4N}({\bf 1}_{N \times L}-{\bf O})^T({\bf 1}_{N \times L}-{\bf O})\\
                = &\frac{1}{4N} [N{\bf 1}_{L \times L} - {\bf 1}_{L\times N}{\bf O} - ({\bf 1}_{L\times N}{\bf O})^T + {\bf O}^T{\bf O}].
\end{array}
\end{equation}

\section{Learning To Diversify}
\label{sec:l2d}
In diversity-based ensembles, the diversity is measured on the classifiers' outputs.
We argue that by introducing kernels, i.e., classifiers' outputs are kernelled, the diversity measure via kerneling can be adaptive and more useful in a variety of situations for different data.
Moreover, we are also sure that by introducing weights, i.e., classifier's outputs are weighted, the diversity measure via weighting can be more representative for various data with noise and redundancy in diversity-based ensembles.
Consequently, we expand the ensemble diversity via weighted kernels, and propose a novel ensemble method, learning to diversity via weighted kernels. We hope that this new method can learn to adaptively select and combine a subset of classifiers by considering both accuracy and (adaptive) diversity.

\subsection{Learning to Diversify via Weighted Kernels}

We first summarize our Learning TO Diversify via Weighted Kernels (\textbf{L2DWK}) approach by giving a direct but simple optimization criterion.
Given a response for the truth of the classifiers' outputs $Y=\{y_1,y_2,...,x_N\}$, classifier weights ${\bf w}$, a kernel $\Bbbk$ for a classifier'output, and the kernel weights $\alpha$, we optimize the following criterion,
\begin{equation}\label{eq:UnifiedCriterion}
\begin{array}{rl}
&min_{{\bf w}, \alpha} loss(Y, {\Bbbk}_\alpha, {\bf w}) - \lambda div(Y, {\Bbbk}_\alpha, {\bf w}),\\
&s.t. ~{\bf w} \succeq 0, ~{\bf 1}^T{\bf w}=1.\\
\end{array}
\end{equation}
where $loss()$ is the loss function of the classification error, $div()$ is the diversity of the ensemble,
and $\lambda$ is the diversity regularization control parameter.
Obviously, this criterion maximizes the accuracy and diversity of the ensemble simultaneously.
Thus, learning to diversify is integrated into the original combination problem formulation
through the accuracy and diversity loss function with weighted kernels. Selecting component classifiers
and measuring the diversity are achieved automatically by optimizing this criterion.

We then present analysis and formulations of \textbf{L2DWK} in detail.
For classifiers $h_i$ and $h_j$, the products of ${\bf 1}_{N\times 1}^T$ and ${\bf O}$, and ${\bf O}^T$ and ${\bf O}$ are kernelled as
$$
{\Bbbk}({\bf 1}_{N\times 1}^T{\bf O})_j = \Bbbk({\bf 1}_{N\times 1},{\bf O}_j);
$$
$$
{\Bbbk}({\bf O}^T{\bf O})_{i,j} = \Bbbk({\bf O}_i,{\bf O}_j).
$$
Next, these kernels are weighted, and we have
\begin{equation}\label{eq:weighted_kernel}
\begin{array}{rl}
{\Bbbk}_\alpha({\bf 1}_{N\times 1}^T{\bf O})_j = \sum_k\alpha_k\Bbbk(1,{\bf O}_{kj}), \\
{\Bbbk}_\alpha({\bf O}^T{\bf O})_{i,j} = \sum_k\alpha_k\Bbbk({\bf O}_{ki},{\bf O}_{kj}), \\
\end{array}
\end{equation}
where $k=1,2,...,N$ for data samples.

Now if we use the average accuracy $\bf P$ (Equation~\eqref{eq:Accuracy}) to compute the classification (error) loss, we have
$$
loss(Y, {\bf w}) = - {\bf 1}_{N \times 1}^T{\bf O}{\bf w}.
$$
With the weighted kernel ${\Bbbk}_\alpha$, there is
\begin{equation}\label{eq:accuracy_loss}
loss(Y, {\Bbbk}_\alpha, {\bf w}) = -{\Bbbk}_\alpha({\bf 1}_{N \times 1}^T{\bf O}){\bf w}.
\end{equation}
Using the pairwise diversity for $div()$ (Equation~\eqref{eq:diversity-1}), we have
$$
div(Y, {\bf w}) = {\bf w}^Tf_D({\bf O}^T{\bf O},{\bf 1}^T{\bf O}){\bf w}
$$
Embedding the weighted kernel ${\Bbbk}_\alpha$, we will get
\begin{equation}\label{eq:div_loss}
div(Y, {\Bbbk}_\alpha, {\bf w}) = {\bf w}^Tf_D({\Bbbk}_\alpha({\bf O}^T{\bf O}),{\Bbbk}_\alpha({\bf 1}^T{\bf O})){\bf w}
\end{equation}
Consequently, by combining the above two formulations (Equation~\eqref{eq:accuracy_loss} and~\eqref{eq:div_loss}), the criterion (in Equation~\eqref{eq:UnifiedCriterion}) changes to
\begin{equation}\label{eq:UnifiedCriterion2}
\footnotesize
\begin{array}{rl}
&min_{{\bf w}, \alpha} loss(Y, {\Bbbk}_\alpha, {\bf w}) - \lambda div(Y, {\Bbbk}_\alpha, {\bf w}) = \\
&min_{{\bf w}, \alpha} - {\Bbbk}_\alpha({\bf 1}^T{\bf O}){\bf w} - \lambda {\bf w}^Tf_D({\Bbbk}_\alpha({\bf O}^T{\bf O}),{\Bbbk}_\alpha({\bf 1}^T{\bf O})){\bf w}\\
\end{array}
\end{equation}

If we use the \emph{Disagreement} diversity (Equation~\eqref{eq:Dis}), the element of ${{\bf D}_{{\Bbbk}_\alpha}}$  will equal
\begin{equation}\label{eq:Dis-Omega}
\footnotesize
  {\bf D}_{\texttt{dis},{\Bbbk}_\alpha}  = f_{D_{\texttt{dis}}}({\Bbbk}_\alpha({\bf O}^T{\bf O}),{\Bbbk}_\alpha({\bf 1}^T{\bf O}))  =  \frac{1}{2}({\bf 1}_{L \times L} - {\Bbbk}_\alpha({\bf O}^T{\bf O}))
\end{equation}
Replacing Equation~\eqref{eq:Dis-Omega} in Equation~\eqref{eq:UnifiedCriterion2}, the optimization function is converted to
\begin{equation}\label{eq:UnifiedCriterion3}
\begin{array}{rl}
min_{{\bf w}, \alpha} - {\Bbbk}_\alpha({\bf 1}^T{\bf O}){\bf w} - \lambda {\bf w}^T \frac{1}{2}({\bf 1}_{L \times L} - {\Bbbk}_\alpha({\bf O}^T{\bf O})) {\bf w}\\
\end{array}
\end{equation}
Similarly, with the \emph{Double Fault} diversity (Equation~\eqref{eq:Df}), the element of ${{\bf D}_{{\Bbbk}_\alpha}}$ is equal to
\begin{equation}\label{eq:Df-Omega}
\footnotesize
  \begin{array}{rl}
  & {\bf D}_{\texttt{df},{\Bbbk}_\alpha}  = f_{D_{\texttt{df}}}({\Bbbk}_\alpha({\bf O}^T{\bf O}),{\Bbbk}_\alpha({\bf 1}^T{\bf O})) = \\
  & \frac{1}{4N} [{\bf 1}_{L \times L} - {\Bbbk}_\alpha({\bf 1}_{L\times N}{\bf O}) - {\Bbbk}_\alpha({\bf 1}_{L\times N}{\bf O})^T + {\Bbbk}_\alpha({\bf O}^T{\bf O})] \\
  \end{array}
\end{equation}
Replacing Equation~\eqref{eq:Df-Omega} in Equation~\eqref{eq:UnifiedCriterion2}, the optimization function changes to
\begin{equation}\label{eq:UnifiedCriterion4}
\footnotesize
\begin{array}{rl}
& min_{{\bf w}, \alpha} - {\Bbbk}_\alpha({\bf 1}^T{\bf O}){\bf w} - \\
& \lambda {\bf w}^T \frac{1}{4N} [{\bf 1}_{L \times L} - {\Bbbk}_\alpha({\bf 1}_{L\times N}{\bf O}) - {\Bbbk}_\alpha({\bf 1}_{L\times N}{\bf O})^T + {\Bbbk}_\alpha({\bf O}^T{\bf O})] {\bf w}\\
\end{array}
\end{equation}
As a result, given a general kernel $\Bbbk$ (linear, Gaussian or polynomial) and the kernel weights $\alpha$, the object optimization (Equation~\eqref{eq:UnifiedCriterion}) in our \textbf{L2DWK} approach can be conveniently converted to a convex quadratic programming problem (Equation~\eqref{eq:UnifiedCriterion3} or~\eqref{eq:UnifiedCriterion4}).

We finally present some remarks about L2DWK.
 If we use the average accuracy to calculate $loss()$ and the pairwise diversity to calculate $div()$ but without weighted kernels, then our \textbf{L2DWK} model (Equation ~\eqref{eq:UnifiedCriterion}) changes to
\begin{equation}\label{eq:QPD}
  \begin{array}{rl}
  & {\bf w_{opt}} = argmin_{\bf w} -\lambda {\bf w}^T{\bf D}{\bf w} - {\bf P}{\bf w}\\
  s.t. &~ {\bf w_{opt}} \succeq 0, ~{\bf 1^Tw_{opt}}=1.
  \end{array}
\end{equation}
Classifier weights of the ensemble can be optimized by solving Equation ~\eqref{eq:QPD}. We call such convex optimization as \textbf{Q}uadratic \textbf{P}rogramming problem with \textbf{D}iversity (QPD) for classifier ensemble, the similar idea of which can be seen in~\cite{Li09}.


\subsection{Self-Training Algorithm}
Generally, it is difficult to find the solution for the optimization in Equation \eqref{eq:UnifiedCriterion2} without known both ${\bf w}$ (classifier weights) and ($\alpha$/$\Bbbk$) (kernel weights).
However, with known $\Bbbk$ and $\alpha$, the optimization is simplified to a convex quadratic programming problem for learning classifiers' weights.
Consequently, we propose a self-training algorithm (shown in Algorithm I) for \textbf{L2DWK}.

\begin{table*}
\label{alg:l2dwk}
\begin{center}
\small
\begin{tabular}{|l|} 
\hline
  Algorithm I: Self-training algorithm for \textbf{L}earning \textbf{TO} \textbf{D}iversify via \textbf{W}eighted \textbf{K}ernels (\textbf{L2DWK})\\
\hline
  \textbf{Input:}\\
  ~~~~~~~$Tr$: the training set. $|Tr|=N$\\
  ~~~~~~~$H=\{h_1,h_2,...,h_L\}$: the base classifier set, $|H|=L$.\\
  ~~~~~~~$\mathbf{D}$: pairwise diversity.\\
  ~~~~~~~$\Bbbk$: a kernel function.\\
  \textbf{Output:} \\
  ~~~~~~~${\bf w}$: the classifier weights.\\
  \textbf{Parameters:}\\
  ~~~~~~~$T$: the max iteration. \\
  ~~~~~~~$\alpha^t$: a $1\times N$ vector (the kernel weights), and $\alpha_i^t$ is the kernel weight of sample $x_i$ at the $t^{th}$ iteration. \\
  ~~~~~~~$\alpha^{t\ast}$: a $1\times N$ vector (the new kernel weights for updating) at  the $t^{th}$ iteration. \\
  ~~~~~~~$\epsilon_t$: the ensemble error rate at the $t^{th}$ iteration. \\
  ~~~~~~~$\beta_t$: a parameter that $\beta_t \in [0,1]$, and $\beta_t \leq \beta_{t+1}$. \\
  \textbf{Procedure:}\\
  ~1:~~~~~~Set $\alpha^{1}_{i}=1/N$. \\
  ~2:~~\textbf{For} $t=1,2,...,T$\\
  ~3:~~~~~~Learn classifier weights {\bf w} (Equation~\eqref{eq:UnifiedCriterion2}) with weighted kernels ($\Bbbk$ and $\alpha^{t}$).\\
  ~4:~~~~~~Calculate the ensemble classification error $\epsilon_t$ by ${\bf w}$ and $Tr$.\\
  ~5:~~~~~~Compute new kernel weights $\alpha^{t\ast}$ with $\epsilon_t$ (Equation~\eqref{eq:L2DWK-Hinge} or \eqref{eq:L2DWK-Exp}).\\
  ~6:~~~~~~Update kernel weights $\alpha^{t+1}$ with $\alpha^{t}$ and $\alpha^{t\ast}$ (Equation~\eqref{eq:updating}).\\ 
  ~7:~~\textbf{End}\\
  \hline
\end{tabular}
\end{center}
\end{table*}

There are three important steps in this learning algorithm: initialization, classifier weight calculation, and kernel weight updating, which are sequentially presented in the following subsections in detail.
At last, we also empirically analyze the convergence performance of this self-training algorithm.

\subsubsection{Initialization}
In the self-training algorithm, firstly given the training set $Tr$ and base classifiers $H$ for optimization, assign the pairwise diversity $\mathbf{D}$.
Then set the max epoch $T$ as a stop constraint, and $\alpha_i=1/N$ for each sample $x_i$, where $N$ is the number of samples in $Tr$.
In our method, we can use general kernels ($\Bbbk$), e.g., linear, polynomial, and Gaussian kernels.

\subsubsection{Classifier Weight Calculation}
In each iteration, given a kernel with known weights $\alpha$, we first use Equation \eqref{eq:UnifiedCriterion2} and Equation \eqref{eq:weighted_kernel} to calculate the base classifier weights vector ${\bf w}$.
As described above, this optimization can be solved as a typical convex quadratic programming problem. Then samples of training set $Tr$ are classified with ${\bf w}$, and the ensemble classification error rate $\epsilon_t$ is calculated.

\subsubsection{Kernel Weight Updating}
Updating the kernel weights $\alpha^t$ is a key process in \textbf{L2DWK}.
In the self-training algorithm, we assume the kernel weights $\alpha^{t+1}$ have a relationship with $\alpha^{t}$, and use a dynamically damped trick, i.e., the damped factor $\beta_t \in [0,1]$ and $\beta_t \leq \beta_{t+1}$. We set $\beta_t$ at the $t^{th}$ iteration as
\begin{equation}\label{eq:beta}
  {\beta_t}    =  \frac{1}{t}.
\end{equation}
Then, we use the following equation to update kernel weights for the $(t+1)^{th}$ iteration,
\begin{equation}\label{eq:updating}
\alpha^{t+1}=\beta_t\alpha^{t\ast} +(1-\beta_t)\alpha^{t},
\end{equation}
where $\alpha^{t}$ and $\alpha^{t\ast}$ are the original and new kernel weights at the $t^{th}$ iteration respectively.

Here, we want the new weight vector $\alpha^{t\ast}$ for kernel weight updating to increase the weights of easily wrong-classified samples.
Correspondingly, we design two methods to calculate $\alpha^{t\ast}$ for \textbf{L2DWK}.

One method, called \textbf{L2DWK}-Hinge, gets the idea from the hinge loss~\cite{Rosa04}. The hinge loss is a loss function used to train classifiers in conventional machine learning techniques. It is usually used for ``maximum-margin" classification, e.g., Support Vector Machines. \textbf{L2DWK}-Hinge computes $\alpha^{t\ast}$ as
\begin{equation}\label{eq:L2DWK-Hinge}
  \alpha^{t\ast}_i=\left\{\begin{array}{rl}
  \frac{1}{N{\displaystyle \epsilon_t}} & m_i \leq 0\\
  0         & otherwise\\
  \end{array}
  \right.,
\end{equation}
where $m_i$ is the margin of sample $x_i$ (Equation~\eqref{eq:Margin}),  $N\epsilon_t$ is the number of samples which are wrongly classified by the ensemble at iteration $t$.

The other method, called \textbf{L2DWK}-Exp, gets the idea from the adaptive re-weighting step in Boosting~\cite{Freund96}.
In Boosting, a distribution of weights over training samples is adaptively maintained, and base classifiers are created sequentially with each classifier concentrating on instances that are not well learnt by previous ones.
Similarly, \textbf{L2DWK}-Exp updates $\alpha^{t\ast}$ by
\begin{equation}\label{eq:L2DWK-Exp}
  \begin{array}{rl}
  & \theta = \frac{1}{2}ln\frac{\displaystyle 1-\epsilon_t}{\displaystyle \epsilon_t},\\
  & \alpha_i^{t+1,\ast} = \frac{\displaystyle \alpha_i^{t\ast} exp(-\theta m_i)}{\displaystyle Z_{t+1}}, \\
  \end{array}
\end{equation}
where $m_i$ is the margin of sample $x_i$ (Equation~\eqref{eq:Margin}), $Z_{t+1}$ is a normalization factor so that $\alpha^{t+1,\ast}$ is still a valid distribution.

In addition, for both above methods, as $\sum_{i=1}^N\alpha_i^{1}=1$, the sum of vector $\alpha^{t+1}$ is
\begin{equation}\label{eq:DSWL-tr}
\footnotesize
  \begin{array}{rl}
  \sum_{i=1}^N\alpha^{t+1}  =& \sum_{i=1}^N(\beta_t\alpha_i^{t\ast} +(1-\beta_t)\alpha_i^{t})\\
                            =& \beta_t \sum_{i=1}^N(\alpha_i^{t\ast}) +(1-\beta_t)\sum_{i=1}^N(\alpha_i^{t}) \\
                            =& \beta_t + (1-\beta_t)\sum_{i=1}^N(\alpha_i^{t}) \\
                            =& \beta_t + (1-\beta_t)\beta_{t-1} + \ldots \prod_{p=1}^t(1-\beta_p)\sum_{i=1}^N(\alpha_i^{1}) \\
                            =& \beta_t + (1-\beta_t)\beta_{t-1} + \ldots \prod_{p=1}^t(1-\beta_p) \\
                            =& 1 \\
  \end{array}
\end{equation}
Obviously, there are $~\sum_{i=1}^N\alpha_i^{t}=1~(\forall t)$ and $\alpha_i^t \in [0,1]~(\forall~t,i)$.

\section{Experiments}
\label{sec:exp}
In this section, we first extensively compare our L2DWK methods with several state-of-the-art methods on a variety of $32$ UCI classification datasets. 
Then, for parameter selection in L2DWK, we perform some experiments with different values of the diversity control parameter $\lambda$ and different kernels. Next, we present experiments on ensembles with different regularization components (e.g, accuracy, or/and diversity with or without weighted kernels). Finally, we empirically analyze the relation between accuracy and diversity in ensembles.

\subsection{Experimental Setting and Datasets}

Our L2DWK method is performed with three common kernels, i.e., linear (L2DWK-Linear), Gaussian (L2DWK-Gauss) and Polynomial kernels (L2DWK-Poly). We use base classifiers generated from two baseline ensembles, Bagging and Random Forests. Please note that without specification, in the following experiments the Classification and Regression Tree (CART) classifier is used as the base learner in all ensembles (see discussions in Section~\ref{sec:regcom}), where CART in Bagging and AdaBoost is derived from the standard Matlab CART function and CART in Random Forests is from a Random Forests toolbox~\footnote{\url{https://code.google.com/p/randomforest-matlab/}.}.
The \emph{Disagreement} (Equation~\eqref{eq:Dis}) is chosen to measure the ensemble diversity.
The L2DWK-Hinge (Equation~\eqref{eq:L2DWK-Hinge})~\footnote{We perform experiments of L2DWK methods with both L2DWK-Hinge (Equation~\eqref{eq:L2DWK-Hinge}) and L2DWK-Exp (Equation~\eqref{eq:L2DWK-Exp}). As the experimental results are very similar, we only present results with L2DWK-Hinge in this paper.} is used to compute and update kernel weights in L2DWK methods.
  Moreover, $10$-fold cross validation is conducted on each dataset, and the validation set is bootstrapped from the training set. A plenty number of base classifiers (here, $301$)~\footnote{The number of base classifiers for Bagging is always default $101$, while the number for Random Forests is sometimes default $501$. In our experiments, we select $301$, a middle value, for both Bagging and Random Forests.} are trained in Bagging or Random Forests for all experimental ensembles.

To verify the effect of the proposed method compared to a number of other ensembles, $32$ typical datasets from UCI machine learning repository~\cite{UCI} are used in our experiments. These datasets are rather challenging with a low accuracy for general classification techniques (e.g., CART). More information for the datasets is presented in Table~\ref{tab:Ids}.

\begin{table}[htb!]
\centering
\caption{\footnotesize{Information for $32$ UCI datasets.}}
\label{tab:Ids}
\scriptsize
\begin{tabular}{|c|ccc|}
  \hline
  Dataset & Instances    & Attributes & Classes \\
  \hline
  artificial	& $5109$	& $7$	& $10$	\\
audiology	& $226$	& $69$	& $24$	\\\hline
auto-mpg	& $399$	& $7$	& $4$ \\
autos	& $205$	& $25$	& $7$	\\ \hline
balance-scale	& $626$	& $4$	& $3$	\\
balloons	& $76$	& $4$	& $2$	\\ \hline
breast-cancer	& $286$	& $9$	& $2$	\\
bridges2	& $108$	& $11$	& $6$	\\\hline
clean1	& $476$	& $166$	& $2$	\\
colic	& $368$	& $22$	& $2$	\\\hline
credit-a	& $695$	& $15$	& $2$	\\
diabetes	& $768$	& $8$	& $2$	\\\hline
echocardiogram	& $132$	& $8$	& $2$	\\
flag	& $194$	& $27$	& $6$	\\\hline
german	& $1000$	& $24$	& $2$	\\
glass	& $214$	& $9$	& $7$	\\\hline
hayes-roth	& $132$	& $4$	& $3$	\\
heart-c	& $303$	& $13$	& $5$	\\\hline
heart-h	& $294$	& $13$	& $5$	\\
heart-statlog	& $270$	& $13$	& $2$	\\\hline
hepatitis	& $155$	& $19$	& $2$	\\
kr-vs-kp	& $3196$	& $36$	& $2$	\\\hline
led24	& $3200$	& $24$	& $10$	\\
led7	& $3200$	& $7$	& $10$	\\\hline
lymph	& $148$	& $18$	& $4$	\\
machine	& $209$	& $7$	& $8$	\\\hline
sonar	& $208$	& $60$	& $2$	\\
vehicle	& $848$	& $18$	& $4$	\\\hline
vowel	& $990$	& $13$	& $11$	\\
wave21	& $5000$	& $21$	& $3$	\\\hline
wave40	& $5000$	& $40$	& $3$	\\
zoo	& $104$	& $17$	& $7$	\\\hline
\end{tabular}
\end{table}


\subsection{Experiments with State-of-the-Art Ensembles}
\label{sec:exp-state}
(Coming soon.)

\section{Conclusions}
\label{sec:conclusion}
Classifier ensemble is widely considered as an effective method to improve accuracy of base classifiers,
which has a variety of applications in pattern recognition, information retrieval and data mining.
Generally speaking, ensemble of diverse classifiers should allow us to get higher accuracy.
However, there is not a definitive connection between diversity measure and ensemble accuracy.
In this paper, we introduce learning to diversify in classifier ensemble, and construct a fairly clear relation between diversity and accuracy.
Specifically, within a linear classifier ensemble framework, we propose Learning TO Diversify via Weighted Kernels (\textbf{L2DWK})
which learns classifier weights by maximizing the accuracy and the diversity simultaneously in a convex minimization problem, where
the diversity is measured with a set of weights on the samples'outputs within a kernel.
Moreover, we propose a self-training algorithm to adaptively learn classifier weights and kernel weights.

\ifCLASSOPTIONcompsoc
  \section*{Acknowledgments}
\else
  \section*{Acknowledgment}
\fi


\ifCLASSOPTIONcaptionsoff
  \newpage
\fi



\bibliographystyle{IEEEtran}
\bibliography{IEEEabrv,diversity}
\end{document}